\documentclass[11pt]{article}
\usepackage[final]{acl}
\usepackage{times}
\usepackage{latexsym}
\usepackage{comment}
\usepackage{tabularx}
\usepackage{array}
\usepackage{makecell}
\usepackage{booktabs}
\usepackage{bm}
\usepackage{tikz}
\usepackage{float}
\usetikzlibrary{arrows.meta,positioning, calc}
\usepackage{float}
\usepackage{wrapfig}
\usepackage{wrapfig}
\usepackage{pgfplots}
\pgfplotsset{compat=1.18}
\usepackage{pgfplotstable}
\usepackage[most]{tcolorbox}
\usepackage[table]{xcolor}
\usepackage{pgfplotstable}
\usepackage{amssymb}
\usepackage{multirow}
\usepackage{pifont}
\usepackage{longtable}
\usepackage{subcaption}
\usepgfplotslibrary{groupplots}
\usepackage[T1]{fontenc}
\usepackage[utf8]{inputenc}
\usepackage{microtype}
\usepackage{inconsolata}
\usepackage{graphicx}
\usepackage[final]{acl}

\title{Evaluating Feedback Focus and Pedagogical 
Adaptivity in LLM-Generated Feedback on Student Writing}

\author{
{\bf Norah Almousa, Shayan Peyghambari Oskoui,} \\
{\bf Raquel Coelho, Gayle Rogers,} \\
{\bf Xiang Lorraine Li, Diane Litman} \\
University of Pittsburgh \\
\{\texttt{nia135, dlitman}\}\texttt{@pitt.edu}
}

\begin{document}
\maketitle
\begin{abstract}
We investigate whether state-of-the-art large language models (LLMs) generate feedback that reflects the pedagogical practices of expert teachers in terms of feedback focus and adaptivity. Previous evaluation efforts have examined feedback characteristics, its impact on learning, and its target, yet the focus of feedback and its adaptivity remains largely overlooked. To bridge this gap, we adopt and refine Narciss's taxonomy into seven feedback focus types to annotate teacher and LLM-generated feedback across three university writing courses. We release \textsc{FeedType}, a benchmark containing annotated teacher and LLM feedback from six LLMs under three prompting strategies. We assess the coverage and distribution of feedback focus types, and examine whether LLMs adapt their feedback across draft stages and student performance levels as an expert instructor does. Our findings show that while most LLMs cover most feedback focus types, they fail to reflect teacher feedback distributions and show varying levels of adaptivity, with none matching the teachers' adaptive behavior. We believe \textsc{FeedType} will support future research on pedagogical alignment in LLM  feedback generation. 

\end{abstract}

\section{Introduction}

Large language models (LLMs) have emerged as promising tools for providing personalized feedback on student essays at scale, with recent work demonstrating success across creative 
\cite{rashkin2025help}, second language 
\cite{wang2025llms}, and academic 
\cite{steiss2024comparing} writing. However, research across multiple educational tasks, including tutoring \cite{tack2022ai}, math problem solving \cite{macina2023mathdial} and essay feedback \cite{dai2023can} has shown that LLMs fail to capture the pedagogical practices 
expert teachers use to guide student learning. For example, in tutoring, \citet{macina2023mathdial} found that ChatGPT tends to provide direct answers, 
while expert teachers guide students through hints. 
This has motivated research on \textit{pedagogical alignment} \cite{macina2023mathdial,macina2025mathtutorbench,sonkar2024pedagogical,maurya2025unifying}, which measures how well LLM behavior aligns with expert teaching 
\cite{perczel2025teachlm}.

Recently, a growing body of work has begun incorporating pedagogical dimensions into LLM-generated feedback evaluation, drawing from established feedback models that characterize effective teacher feedback \cite{hattie2007power,yang2013feedback}. For example, \citet{sessler2025towards} and \citet{dai2023can} applied \citet{hattie2007power}'s framework,  examining feedback's learning supportive components and who feedback targets. \citet{mah2025sentence} applied \citet{yang2013feedback}'s framework, focusing on how feedback is delivered across cognitive, social, and structural dimensions. 
However, 
\textit{feedback focus} -- what teachers focus on when responding to student writing \cite{yu2014analysis,yu2014understanding,lipnevich2021review}, remains unexplored. This is important since teachers vary their feedback focus to address different aspects of student writing \cite{lipnevich2021review}, yet LLMs have shown limited ability to identify writing issues \cite{rashkin2025help,sessler2025towards}, motivating us to examine what feedback focus types LLMs tend to provide and how this compares to expert teachers.

The feedback literature has classified feedback focus into different levels, ranging from broad distinctions such as content versus form \cite{ferris1997influence} to more detailed taxonomies distinguishing specific types such as mistakes, tasks, and concepts \cite{narciss2008feedback}. Recognizing multiple feedback focus types is pedagogically important, as it informs what to address, for whom, and when. Expert teachers leverage this knowledge by shifting their focus across draft stages and adapting to each student's individual needs \cite{ortmeier2006dana,ferris1997influence}, reflecting a deeply adaptive teaching practice. To our knowledge, feedback focus has not been examined in the evaluation of LLM-generated feedback, specifically in terms of what aspects LLMs tend to focus on and whether they exhibit the same adaptive behavior across draft stages and student performance levels. 
We aim to bridge this gap by addressing three research questions: \\[0.1em]
\noindent\textbf{RQ1:} {\it What feedback focus types do LLMs tend to generate, and how does this compare to expert teachers?}\\[0.1em]
\noindent\textbf{RQ2:} {\it Do LLMs adapt their feedback focus across draft stages, as an expert teacher does?}\\[0.1em]
\noindent\textbf{RQ3:} {\it Do LLMs adapt their feedback focus based on student performance, as an expert teacher does?}

To address these questions, we present \textsc{FeedType}\footnote{Data 
available at \url{https://github.com/nia135/FeedType}.}, a dataset of expert teacher feedback on student essays across three university writing courses (taken from the publicly available SEFORA corpus \cite{oskoui2026sefora}), with teacher and LLM-generated feedback annotated for feedback focus. 
While SEFORA provides rich teacher feedback data, it lacks feedback focus annotations needed to address our research questions. To bridge this gap, we adapt an existing taxonomy \cite{narciss2008feedback} and annotate all feedback comments in \textsc{FeedType} according to feedback focus. We then prompt six LLMs, covering open and closed weight 
models as well as the LearnLM specialized pedagogical model \cite{team2024learnlm}, to generate feedback on selected paragraphs from \textsc{FeedType} using three prompting strategies. We annotate the LLM generated feedback using the same taxonomy applied to teacher feedback, enabling a direct comparison. \textsc{FeedType} thus serves as a benchmark for analyzing LLM pedagogical practices in terms of feedback focus and whether it adapts to student performance and draft stage. Our findings indicate that while most LLMs cover the majority of feedback focus types, they fail to reflect teacher feedback distributions and show limited adaptivity across draft stages and student performance levels. 


\section{Related Work}
\label{sec:Related Work}
\begin{table*}[t]
\centering
\scriptsize
\setlength{\tabcolsep}{4pt}
\renewcommand{\arraystretch}{1.15}
\begin{tabular}{p{2.8cm} p{1cm} p{1.8cm} p{4.5cm} p{1.3cm} p{1.1cm} p{1.2cm}}
\toprule
\textbf{Corpus} & \textbf{Domain} & \textbf{Feedback Granularity} & \textbf{What Annotated} & \textbf{Feedback} & \textbf{Scores} & \textbf{Revisions} \\
\midrule
\multicolumn{7}{l}{\textbf{L2 Learner Writing}} \\
\midrule
TOEFL11 \cite{blanchard2013toefl11}
& L2 & Essay
& Holistic English proficiency scores (low, medium, high)
& No & Yes & No \\
CLC-FCE \cite{yannakoudakis-etal-2011-new}
& L2 & Sentence
& Grammatical error annotations and corrections
& No & Yes & No \\
EFCAMDAT \cite{geertzen2013automatic}
& L2 & Essay
& English proficiency levels and learner metadata
& No & Yes & Yes \\
Write \& Improve \cite{nguyen-etal-2018-empirical}
& L2 & Essay
& Grammatical error corrections, CEFR proficiency levels, and learner revisions
& Automated & Yes & Yes \\
LEAF \cite{behzad-etal-2024-leaf}
& L2 & Essay
& Personalized holistic writing feedback and revision annotations
& Teacher \& Automated & No & Yes \\
ICNALE \cite{Ishikawa}
& L2 & Essay + Sentence
& Proficiency labels and grammatical error correction annotations
& No & Yes & No \\
FCG Corpus \cite{nagata-etal-2020-creating}
& L2 & Sentence
& Feedback comments explaining grammatical errors and writing rules
& Teacher & No & No \\
\midrule
\multicolumn{7}{l}{\textbf{Academic Writing}} \\
\midrule
ASAP++ \cite{mathias-bhattacharyya-2018-asap}
& Native & Essay
& Holistic and trait-specific essay scores (content, organization, word choice, sentence fluency, etc.)
& No & Yes & No \\
ASAP 2.0 \cite{crossley2025large}
& Native & Essay
& Holistic essay scores for source-based argumentative essays
& No & Yes & No \\
\midrule
\multicolumn{7}{l}{\textbf{Creative Writing}} \\
\midrule
STORYFEEDBACK \cite{rashkin2025help}
& Creative & Story
& Generated feedback annotated with human ratings (correctness, specificity, relevance, error detection)
& Automated & No & No \\
SciFi-100 \cite{li2026llm}
& Creative & Story
& Novelty and creativity dimensions of generated stories
& No & No & No \\
Towards a Novel Benchmark \cite{wang2025towards}
& Creative & Story
& Narrative, paragraph, and sentence-level literary quality dimensions
& No & No & No \\
\midrule
\multicolumn{7}{l}{\textbf{Our Corpus}} \\
\midrule
\textsc{FeedType}
& L2,
Native,
Creative & Paragraph
& Feedback with feedback focus type labels
& Teacher \& Automated & Yes & Yes \\
\bottomrule
\end{tabular}
\caption{Overview of writing and feedback corpora across L2 learner, academic, and creative 
writing domains. \textsc{FeedType} is the only one that jointly provides scores, paragraph-level 
feedback, feedback focus type annotations, and revisions across all three domains.}
\label{tab:writing_feedback_corpora}
\end{table*}

\paragraph{LLMs for Feedback Generation on Writing}
LLMs have been used for feedback generation across multiple writing contexts, including L2 learner writing \cite{wang2025llms,kaneko2024controlled,mizumoto2024testing}, creative writing \cite{rashkin2025help}, and academic writing \cite{chu2026feedeval}. Feedback generation has been explored at the sentence level \cite{mizumoto2024testing,song2024gee,kaneko2024controlled,banno2024grammatical} and on complete student essays \cite{chu2026feedeval,behzad2024assessing,rashkin2025help}. However, to our knowledge, feedback generation at the paragraph level has received very little attention, despite evidence that this is a common and valuable form of expert teacher feedback \cite{mah2025sentence,hyland2006feedback,zamel1985responding} that LLMs fail to provide \cite{mah2025sentence}. We address this gap by generating paragraph level feedback across multiple writing domains (L2 learner writing, composition, and fiction), where we provide LLMs with the complete student essay and target specific paragraphs within it for feedback generation.

\paragraph{Evaluation of LLM-Generated Feedback}
Many studies have evaluated LLM-generated feedback in terms of accuracy \cite{yoon2023evaluation}, quality \cite{algobaei2026prompt}, and bias \cite{du2025benchmarking, tan2026marked}. Fewer have examined its alignment with expert teacher practices. In student writing, prior work has compared LLM and teacher generated feedback in terms of quality \cite{steiss2024comparing}, content and targets \cite{dai2023can}, and diversity \cite{tan2026can}. We extend prior work on alignment in two ways. First, we examine feedback focus and its adaptivity, dimensions that have received limited attention in prior evaluations. Second, we compare the distribution of feedback focus types between teachers and LLMs using population-level evaluation approaches adopted in prior educational research \cite{mah2025sentence,kucheria2025comparing,asano2025can}. 


\paragraph{Feedback Focus and Adaptivity}
Prior work has classified feedback focus narrowly, either by grammatical error types or limited broad categories \cite{nagata-etal-2020-creating, steiss2024comparing,liu2025comparative}. To address this gap, we adopt and refine the taxonomy of \citet{narciss2008feedback}, which captures a broader range of feedback focus types observed in expert teacher feedback. This taxonomy allows us to examine how teachers adapt their feedback priorities across draft stages and learner performance levels. For example, early drafts may receive greater attention to content and ideas, whereas later drafts may focus more on vocabulary and style \cite{zamel1985responding}. Similarly, feedback may differ by learner performance level \cite{ferris1997influence}  because students at different levels may benefit from different types of feedback. We refer to these adjustments in feedback focus as feedback adaptivity. To our knowledge, no prior work has examined whether LLMs demonstrate similar adaptive patterns.

\paragraph{Related Corpora}
Across L2, native, and creative writing, existing corpora are limited to scores only without feedback \cite{blanchard2013toefl11, yannakoudakis-etal-2011-new, geertzen2013automatic, Ishikawa, crossley2025large, li2026llm}, sentence-level or holistic error-focused feedback \cite{nagata-etal-2020-creating, nguyen-etal-2018-empirical, behzad-etal-2024-leaf}, multi-dimensional scoring and feedback without feedback focus type annotations \cite{mathias-bhattacharyya-2018-asap}, or essay evaluation without feedback \cite{wang2025towards, rashkin2025help}. To our knowledge, no existing corpus jointly provides scores, revisions, teacher and automatic feedback, and feedback focus type annotations, all of which are needed to analyze feedback focus and its adaptivity across draft stages and student performance levels. We release \textsc{FeedType}, a corpus that provides all of these components, enabling such analysis. We summarize and compare existing datasets in Table~\ref{tab:writing_feedback_corpora}.

\section{\textsc{FeedType} \& Annotation Procedure}

\begin{figure*}
    \centering
    \includegraphics[width=0.9\linewidth]{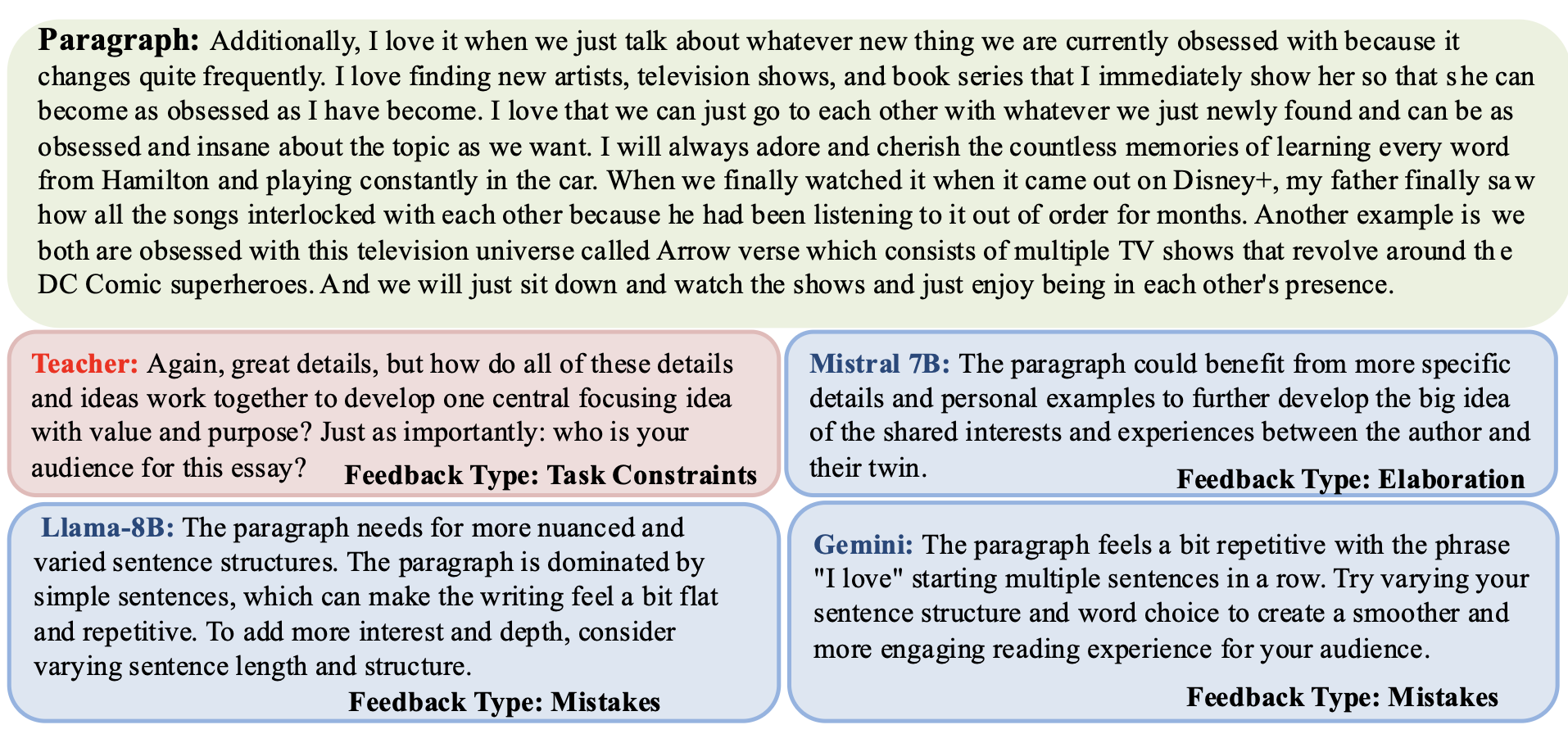}
    \caption{Examples of teacher and LLM feedback 
         on a student paragraph, annotated with 
         feedback types.}
    \label{fig:feedback_example}
\end{figure*}

\paragraph{Construction of \textsc{FeedType}}

We sampled course data from the publicly available SEFORA corpus of first-year college student essays and teacher feedback \cite{oskoui2026sefora}. In particular, we use the data from three different courses (composition, ESL, and fiction) from a single semester, choosing the semester with the highest student enrollment. 
Where a course had multiple classes in the same semester, we included all of them.
The composition course includes two classes taught by the same instructor, while the ESL and fiction courses each consist of one class taught by different instructors. Students in the composition and ESL courses submitted up to three essays with multiple drafts, while fiction students submitted three essays without multiple drafts; in all courses, only the first draft was required. We focus on paragraph-level feedback, where each paragraph receives one teacher comment. Student grades and assignment rubrics are used to support feedback classification and adaptivity analysis. Table~\ref{tab:paragraph_feedback_overview1} summarizes the corpus, and Figure~\ref{fig:feedback_example} shows a feedback example.  Additional corpus details can be found in Appendix \ref{sec:Dataset}.

\begin{table}[H]
    \centering
    \scriptsize
    \setlength{\tabcolsep}{3pt}
    \resizebox{\columnwidth}{!}{%
    \begin{tabular}{llccc}
    \toprule
    \textbf{Course} & \textbf{Essay} & \textbf{Initial} &
    \textbf{Revision} & \textbf{Final} \\
    \midrule
    Composition (Class 1) & 1 & 15/99 & 10/64 & 3/15 \\
    & 2 & 15/82 & 9/55 & 0/0 \\
    & 3 & 13/76 & 0/0 & 8/39 \\
    & \textbf{Total} & 43/257 & 19/119 & 11/54 \\
    \midrule
    Composition (Class 2) & 1 & 15/85 & 13/77 & 2/11 \\
    & 2 & 14/70 & 10/49 & 4/13 \\
    & 3 & 15/78 & 0/0 & 6/21 \\
    & \textbf{Total} & 44/233 & 23/126 & 12/45 \\
    \midrule
    ESL Composition & 1 & 14/148 & -- & 12/38 \\
    & 2 & -- & -- & 11/35 \\
    & 3 & 13/37 & -- & 13/99 \\
    & \textbf{Total} & 27/185 & -- & 36/172 \\
    \midrule
    Introduction to Fiction & 1 & 18/276 & -- & -- \\
    & 2 & 17/183 & -- & -- \\
    & 3 & 18/316 & -- & -- \\
    & \textbf{Total} & 53/775 & -- & -- \\
    \midrule
    \multicolumn{2}{l}{\textbf{Total Paragraphs}} &
    \multicolumn{3}{c}{\textbf{1,966}} \\
    \bottomrule
    \end{tabular}%
    }
    \caption{Essays and selected paragraphs with one feedback (\#Essays / \#Paragraphs).}
    \label{tab:paragraph_feedback_overview1}
\end{table}

\paragraph{Feedback Annotation Scheme}
\label{subsec:Content-Based Annotation Scheme}
We base our annotation scheme on \citet{narciss2008feedback}'s taxonomy, which provides detailed content focus types for feedback analysis. Prior frameworks either conflate content and function \cite{hattie2007power, Kulhavy1989FeedbackIW} or categorize content only broadly \cite{ferris1997influence, yu2014understanding}; Narciss's taxonomy addresses both by treating them separately and offering more precise content focus types. We adapt and extend this taxonomy to better capture the nature of written teacher feedback, retaining relevant feedback types, refining others, and introducing new types to emphasize explanatory feedback. Appendix~\ref{sec:narciss} presents the original and adapted taxonomies. The final taxonomy consists of seven feedback types: 
\textit{Task Constraints}, which clarify assignment 
goals and requirements; \textit{Concepts}, which 
explain key ideas needed to complete the task; 
\textit{Elaboration}, which encourages students to 
expand their ideas with more detail; 
\textit{Clarification}, which asks students to 
clarify unclear parts of their writing; 
\textit{Mistakes}, which identify specific errors 
in the text; \textit{Praise}, which provides 
encouragement or positive evaluation; and 
\textit{Other}, which covers feedback that either does not fit any single focus type or spans multiple types simultaneously. Annotation guidelines 
are provided in Appendix~\ref{sec:annotation-guidelines}, with some annotated feedback examples  in Figure~\ref{fig:feedback_example}.

\paragraph{Annotator Training and Development of Coding Manual}
\label{subsec:AnnotationGuidelines}
Two authors iteratively developed and refined the coding manual through four pilot rounds. In each round, 50 teacher feedback comments  (25 composition, 12 ESL, 13 fiction) and their associated paragraphs were randomly sampled and independently labeled. Disagreements were analyzed, and category definitions and instructions were updated to reduce ambiguity. This process continued until substantial agreement was achieved, with the final round reaching Cohen’s Kappa $\kappa = 0.774$ \cite{sim2005kappa}.

\paragraph{Scaling Annotation}
\label{subsec:Scaling Content-Based Annotation} 
Motivated by prior work demonstrating that LLMs can reliably annotate data at scale \cite{tan2024large}, we use them to automatically classify feedback according to our feedback focus types. We prompted six candidate models to classify the feedback (see Appendix~\ref{sec:prompt-llm-judges} for the prompt) and evaluated their reliability using three methods. We constructed two evaluation datasets serving distinct purposes: (1) 100 randomly sampled feedback instances to measure inter-annotator agreement and assess whether LLM annotations can substitute for human annotators using the Alt-Test \cite{calderon-etal-2025-alternative}, and (2) 400 feedback instances as a test set to evaluate whether conclusions drawn from LLM-generated annotations are consistent with those derived from human labels, ensuring the validity of our annotation pipeline. Gemini 2.0 Flash was selected as the annotator, passing all three evaluations: achieving a Cohen's Kappa of 0.78 with human annotators, passing the Alt-Test, and producing conclusions consistent with those drawn from human annotations. Further details are provided in Appendix~\ref{sec:Evaluating LLMs as Annotators}.

\section{Experiments}
\subsection{Generating Feedback with LLMs}
We generated feedback using six models: \textbf{Mistral-7B-Instruct, Mistral-24B-Instruct} \cite{jiang2023mistral7b}, \textbf{LLaMA-3.1 8B} \cite{meta2024llama3}, \textbf{Qwen2.5-70B-Instruct} \cite{hui2024qwen2}, \textbf{Gemini 2 Flash} \cite{google2025gemini3flash}, and \textbf{LearnLM} \cite{team2024learnlm}, covering both open and closed source instruction tuned models across different model sizes, to examine whether model size and type influence feedback practices. LearnLM is included as a specialized  model fine-tuned for educational tasks \cite{team2024learnlm}. 

Each model was prompted under three conditions: (1) {\textbf{zero-shot} \cite{rashkin2025help,stahl-etal-2024-exploring}, (2) \textbf{zero-shot with the names and definitions of the feedback focus categories}, as providing task specific information has been shown to improve LLM output quality \cite{stahl-etal-2024-exploring}, and (3) {\textbf{few-shot with the names, definitions, and examples of each feedback focus type category}, as in-context examples have been shown to improve LLM performance \cite{brown2020language}. The goal is to examine whether providing feedback type information or examples shifts model behavior toward expert teacher practices. Our prompt follows \citet{rashkin2025help} and \citet{stahl-etal-2024-exploring}, instructing the model to act as a teacher and including the assignment rubric and task prompt. However, we designed the prompt to mirror real teacher context, which is often absent from prior work: (1) the full essay alongside the target paragraph, (2) the draft version, and (3) previously received feedback for revision and final drafts. The full prompt is in Appendix~\ref{sec:prompt-llm-generators}, and Figure~\ref{fig:feedback_example} shows example outputs.


\subsection{Evaluation Methods}
To  examine the pedagogical alignment of LLM feedback focus types  \textbf{(RQ1)}, we compare them to expert teacher feedback  focus types 
in two ways:

\noindent \textbf{(i) Coverage:} How much can LLMs cover teachers' feedback types when responding to student essays? We compute \textbf{Cov\_all}, inspired by \citet{asano2025can}, defined as the number of teacher feedback focus types for which the LLM generates at least one instance, divided by the total number of teacher feedback focus types.

\noindent \textbf{(ii) Similarity:} How similar is the feedback focus type distribution of LLM-generated feedback to that of teachers? We compute \textbf{Jensen-Shannon Divergence (JSD)}, following prior work \cite{lu2020diverging,ma2025large,elangovan2025beyond}. JSD ranges from 0 to 1, where 0 indicates identical  and 1 indicates completely different distributions:
\begin{equation}
\small
JSD(P_{m} \| P_{T}) = \frac{1}{2} D_{KL}(P_{m} \| M) + \frac{1}{2} D_{KL}(P_{T} \| M)
\end{equation}
where $M = \frac{1}{2}(P_{m} + P_{T})$, $P_{m}$ is the distribution of feedback focus types generated by model $m$, $P_{T}$ is the teacher reference distribution, and $D_{KL}$ is the Kullback-Leibler divergence.


To measure whether LLMs adapt their feedback focus types across draft stages as an expert teacher does (\textbf{RQ2}), we follow corpus linguistics practice of using chi-square-based tests to detect categorical distribution changes across conditions \cite{mcenery2011corpus}. Since chi-square does not accept zero values, and some models did not provide feedback on certain focus types, we apply {\textbf{Fisher's Exact Test} \cite{agresti1992survey}, which is statistically equivalent but accepts zero values. For each feedback type $f$, source $s$, and pair of draft stages $(d_i, d_j)$, we construct a $2 \times 2$ contingency table:

\begin{equation}
\begin{pmatrix}
\small
c_{s,f}^{(d_i)} & c_{s,f}^{(d_j)} \\[6pt]
c_{s,\neg f}^{(d_i)} & c_{s,\neg f}^{(d_j)}
\end{pmatrix}
\end{equation}

where $c_{s,f}^{(d)}$ is the count of feedback type $f$ from 
source $s$ at draft stage $d$, and $c_{s,\neg f}^{(d)}$ is the 
count of all other feedback types. The null hypothesis $H_0$ 
states that feedback type $f$ and draft stage are independent, 
in other words, the proportion of $f$ does not change between drafts. We test two pairwise comparisons per feedback type: initial vs.\ revision, and revision vs.\ final. We reject $H_0$ if $p < 0.05$. 


To measure whether  LLMs also adapt their feedback focus based on student performance as an expert teacher does (\textbf{RQ3}), we again 
use \textbf{Fisher's Exact Test} as in RQ2.  However, we now compare the distribution of each feedback focus type between high- and low-performing students. 


\section{Results}
\label{sec:Results}

\begin{figure*}[t]
    \centering
    \includegraphics[width=\textwidth]{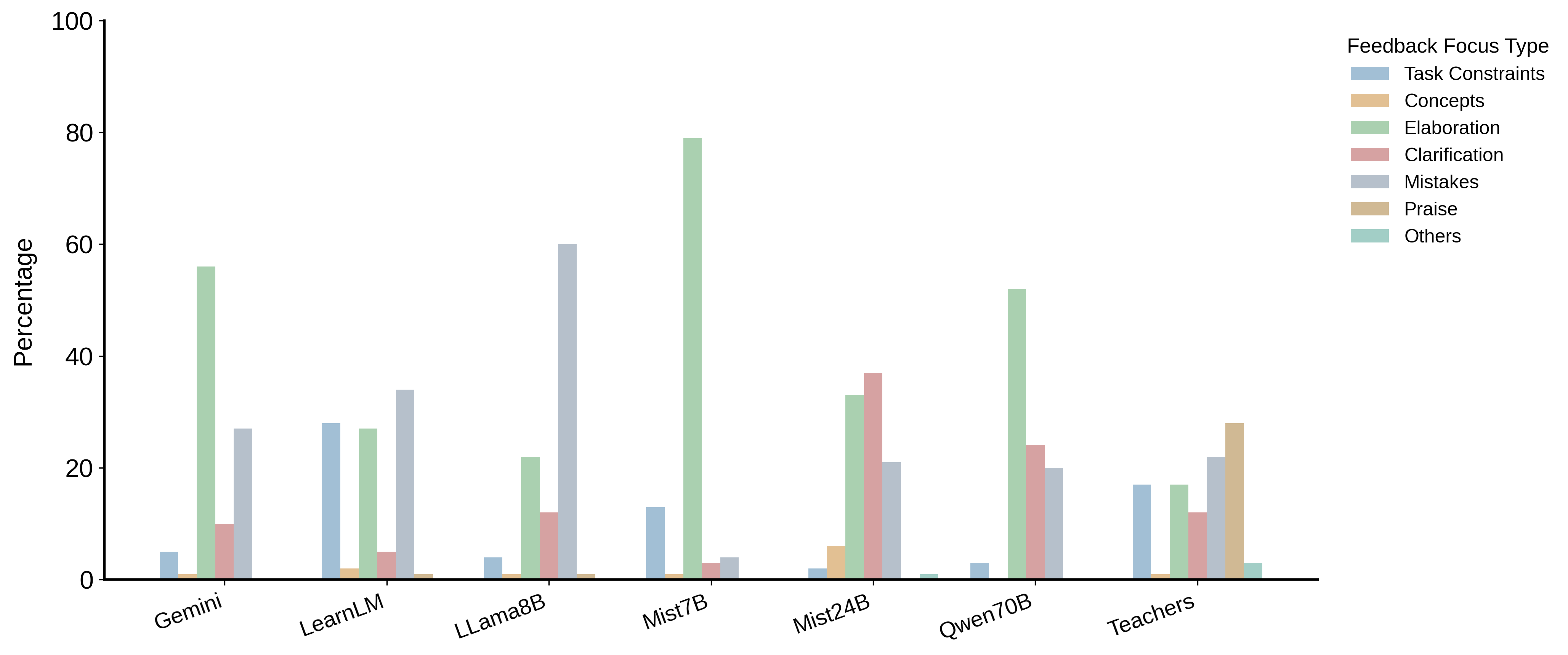}
    \caption{Distribution of feedback focus types (\%) for each model and the teachers under the zero-shot baseline, aggregated across all courses.}
    \label{fig:distribution-feedback-focus}
\end{figure*}

Inspired by prior work on behavioral alignment between LLMs and human tutors \cite{kucheria2025comparing}, we aggregate all teacher feedback to construct a reference distribution, as shown in Figure~\ref{fig:distribution-feedback-focus}.

\textbf{RQ1: What feedback focus types do LLMs tend to generate, and how does this compare to expert teachers?} We address this question in two parts: in terms of coverage and similarity.

\noindent\textbf{(1.1) How much can LLMs cover teachers' feedback types when responding to student essays?} To understand whether each model can generate feedback across different focus types, we use the Cov\_all metric, which measures how many of the seven feedback focus types a model produces. We start by analyzing the zero-shot baseline, where models receive no information about the feedback focus types. As shown in Table~\ref{tab:coverage_missing}, LearnLM, Llama-7B, and Mistral-7B achieve full coverage. However, Gemini and Mistral-24B fail to produce praise-focused feedback, while Qwen-70B does not produce the Other feedback focus type.
A model missing this feedback focus type generates feedback with only one focus per comment, while models that achieve full coverage demonstrate the ability to produce at least one comment that addresses multiple focus types simultaneously, or that is not categorizable as one of the provided focus types. 

To guide the models toward more complete coverage, we modified the prompting by adding information about the feedback focus types and providing few-shot examples. As shown in Table~\ref{tab:coverage_missing}, Mistral-24B benefits from both strategies - producing praise-focused feedback when prompted with either category information or examples, while Qwen-70B just benefits from examples. Gemini, however, still fails to produce praise-focused feedback across both strategies. 
In addition, Gemini shows degradation as it now misses the Other feedback focus type in addition to missing praise. 
Similarly, LearnLM, which achieved full coverage in the zero-shot baseline, begins to miss the Other feedback focus type when provided with category information, indicating that explicit category guidance may shift some models toward generating more single-focused feedback.

\begin{table}[t]
\centering
\scriptsize
\setlength{\tabcolsep}{3pt}  
\begin{tabularx}{\columnwidth}{lXXX}
\toprule
\textbf{Model} & \textbf{Zero-shot} & \textbf{Zero-shot + Categories} & \textbf{Few-shot + Categories} \\
\midrule

Gemini
& \shortstack{0.86 \\ Praise}
& \shortstack{0.71 \\ Praise, Other}
& \shortstack{0.71 \\ Praise, Other} \\
\midrule

LearnLM
& \shortstack{1.0 \\ -}
& \shortstack{0.86 \\ Other}
& \shortstack{0.86 \\ Other} \\
\midrule

Qwen-70B
& \shortstack{0.86 \\ Other}
& \shortstack{0.86 \\ Other}
& \shortstack{1.0 \\ -} \\
\midrule

Llama-7B
& \shortstack{1.0 \\ -}
& \shortstack{1.0 \\ -}
& \shortstack{1.0 \\ -} \\
\midrule

Mistral-7B
& \shortstack{1.0 \\ -}
& \shortstack{1.0 \\ -}
& \shortstack{1.0 \\ -} \\
\midrule

Mistral-24B
& \shortstack{0.86 \\ Praise}
& \shortstack{1.0 \\ -}
& \shortstack{1.0 \\ -} \\
\midrule

\textit{Teacher}
& \multicolumn{3}{c}{\textit{1.0}} \\
\bottomrule
\end{tabularx}

\caption{Coverage of teacher feedback categories. Each cell shows Cov.all (top) and missing feedback types (bottom).}
\label{tab:coverage_missing}
\end{table}

\noindent\textbf{(1.2) How similar is the feedback focus type distribution of LLM-generated feedback to that of teachers?} We examine how similar each model's feedback focus type distribution is to that of teachers using JSD. As shown in Table~\ref{tab:jsd}, in the zero-shot setting, LearnLM achieves the lowest JSD (0.1344), indicating the closest alignment with the teacher distribution, followed by Llama-7B (0.1619) and Mistral-24B (0.1780). Among all models, the educationally fine-tuned LearnLM most closely approximates the teacher feedback focus type distribution, while larger models such as Gemini and Qwen-70B show some of the highest divergence. Applying different prompting strategies yields limited improvement. Prompt 2 increases JSD for most models, while Prompt 3 improves alignment only for Mistral-7B (0.2398). In both cases, the zero-shot baseline remains the closest to the teacher feedback focus type distribution for most models. 

To better understand these results, we examine the feedback focus type distribution more closely. As shown in Figure~\ref{fig:distribution-feedback-focus}, Gemini, Qwen-70B, and Mistral-7B exhibit a strong preference for Elaboration, assigning more than 50\% of their feedback to this type, while Llama-7B prefers Mistakes at over 60\%. This preference for a single type might explain their mostly higher divergence from the teachers, who distribute feedback more evenly. In contrast, LearnLM and Mistral-24B show no such preference, with no single type exceeding 50\%, and are relatively closer to the teacher distribution. Moreover, the strong preference for a single type might 
lead to repetitive, narrow feedback, potentially causing models to overlook other important aspects of student writing. 
\begin{table}[t]
\centering
\scriptsize
\setlength{\tabcolsep}{3pt}
\begin{tabular}{lccc}
\toprule
\textbf{Model} & \textbf{Zero-shot} & \textbf{Zero-shot + Categories} & \textbf{Few-shot + Categories} \\
\midrule
Gemini      & \textbf{0.1816} & 0.1896 & 0.2800 \\
LearnLM     & \cellcolor{gray!20}\textbf{0.1344} & 0.1452 & 0.1486 \\
Qwen-70B    & \textbf{0.1884} & 0.2682 & 0.2376 \\
Llama-7B    & \textbf{0.1619} & 0.2433 & 0.2665 \\
Mistral-7B  & 0.2695 & 0.2454 & \textbf{0.2398} \\
Mistral-24B & \textbf{0.1780} & 0.2236 & 0.1938 \\
\bottomrule
\end{tabular}
\caption{Jensen-Shannon Divergence (JSD) between each model's feedback focus type distribution and the teacher distribution. Lower values indicate closer alignment to the teacher distribution. \textbf{Bold} indicates the lowest JSD per model. \colorbox{gray!20}{Gray} indicates the overall best value.}
\label{tab:jsd}
\end{table}

\textbf{RQ2: Do LLMs adapt their feedback focus across draft stages, as an expert teacher does?}
\label{sec:drafts}
To address this question, we examine whether feedback focus type distributions change across draft versions for both the teacher and each model, using Fisher's Exact Test to detect significant changes, where significance suggests adaptation to the student's draft. This analysis is limited to the composition course, as the fiction course has no drafts and the ESL course was excluded due to limited variation in teacher feedback focus types.

Table~\ref{tab:fisher_drafts} presents zero-shot results, and shows that the teacher demonstrates clear feedback adaptivity across draft stages, with significant changes in five feedback focus types: Task Constraints, Elaboration, Clarification, Mistakes, and Praise, across both stages, from initial to revision and from revision to final. When looking at the models, Mistral-24B shows the closest pattern to the teacher, with significant changes in three feedback focus types: Elaboration, Clarification, and Mistakes, across both stages. Task Constraints, however, shows a significant change only from initial to revision. Llama-7B follows, showing significant changes in two feedback focus types: Task Constraints and Mistakes, across both stages, while Elaboration shows a significant change only from initial to revision. Gemini also shows significant changes in two feedback focus types: Elaboration and Mistakes, across both stages. LearnLM shows significant changes in Mistakes across both stages, while Elaboration changes only from initial to revision. 
Interestingly, all models that demonstrate adaptivity consistently show significant changes in the same two feedback focus types: Mistakes and Elaboration. This aligns with the distribution analysis in Figure~\ref{fig:distribution-feedback-focus}, where these two types appear with the highest proportion across most models. This suggests that models are more capable of adapting feedback focus types they produce frequently.

\begin{table*}[t]
\centering
\scriptsize
\setlength{\tabcolsep}{4pt}
\begin{tabular}{llrrrrrrrc}
\toprule
\textbf{Models/ Teacher} & \textbf{Stage} & 
\textbf{Task Constraints} & 
\textbf{Concepts} & 
\textbf{Elaboration} & 
\textbf{Clarification} & 
\textbf{Mistakes} & 
\textbf{Praise} & 
\textbf{Other} &
\textbf{Sig.} \\
\midrule
\multirow{5}{*}{\textbf{Teacher}}
 & Initial  & 29.3 & 0.8 & 23.8 & 8.3  & 10.7 & 26.9 & 0.2 & \\
 & Revision & 17.2 & 0.0 & 7.5  & 13.4 & 23.8 & 38.1 & 0 & \\
 & Final    & 6.5  & 0.0 & 1.1  & 3.2  & 3.2  & 84.9 &1.1 & \\
 & I vs R   & \textbf{0.0003} & 0.3072 & \textbf{0.0000} & \textbf{0.0371} & \textbf{0.0000} & \textbf{0.0029} & 1.0000 & \textbf{5} \\
 & R vs F   & \textbf{0.0059} & 1.0000 & \textbf{0.0180} & \textbf{0.0031} & \textbf{0.0000} & \textbf{0.0000} & 1.0000 & \textbf{5} \\
\midrule
\multirow{5}{*}{Gemini}
 & Initial  & 10.2 & 0.0 & 53.3 & 9.4  & 27.1 & 0.0 & 0.0& \\
 & Revision & 14.3 & 0.0 & 65.3 & 8.6  & 11.8 & 0.0 & 0.0& \\
 & Final    & 7.1  & 0.0 & 47.5 & 11.1 & 34.3 & 0.0 &0.0 & \\
 & I vs R   & 0.1122 & 1.0000 & \textbf{0.0020} & 0.7866 & \textbf{0.0000} & 1.0000 & 1.0000 & \textbf{2} \\
 & R vs F   & 0.0703 & 1.0000 & \textbf{0.0034} & 0.5386 & \textbf{0.0000} & 1.0000 & 1.0000 & \textbf{2} \\
 \midrule
\multirow{5}{*}{LearnLM}
 & Initial  & 16.5 & 0.2 & 54.1 & 8.2  & 21.0 & 0.0 & 0.0 & \\
 & Revision & 24.9 & 0.0 & 62.9 & 5.7  & 6.5  & 0.0 & 0.0 & \\
 & Final    & 13.1 & 0.0 & 47.5 & 12.1 & 27.3 & 0.0 & 0.0 & \\
 & I vs R   & 0.2350 & 1.0000 & \textbf{0.0000} & 0.2687 & \textbf{0.0001} & 1.0000 & 1.0000 & \textbf{2} \\
 & R vs F   & 0.3891 & 1.0000 & 0.0914 & 0.6942 & \textbf{0.0027} & 1.0000 & 1.0000 & \textbf{1} \\
\midrule

\multirow{5}{*}{LLaMA}
 & Initial  & 43.5 & 0.2 & 24.3 & 3.7  & 28.2 & 0.2 &0.0 & \\
 & Revision & 38.8 & 0.4 & 43.3 & 2.0  & 15.5 & 0.0 & 0.0 & \\
 & Final    & 33.3 & 0.0 & 33.3 & 3.0  & 30.3 & 0.0 & 0.1 & \\
 & I vs R   & \textbf{0.0124} & 1.0000 & \textbf{0.0000} & 0.1454 & \textbf{0.0000} & 1.0000 & 1.0000 & \textbf{3} \\
 & R vs F   & \textbf{0.0240} & 0.2878 & 0.1856 & 0.2842 & \textbf{0.0012} & 1.0000 & 1.0000 & \textbf{2} \\
\midrule
\multirow{5}{*}{Qwen-70B}
 & Initial  & 4.9  & 0.2 & 40.7 & 29.9 & 24.3 & 0.0 &0.0 & \\
 & Revision & 9.0  & 0.0 & 36.7 & 29.4 & 24.9 & 0.0 & 0.0 & \\
 & Final    & 4.0  & 1.0 & 35.4 & 26.3 & 33.3 & 0.0 &0.0 & \\
 & I vs R   & \textbf{0.0360} & 1.0000 & 0.3367 & 0.9319 & 0.8561 & 1.0000 & 1.0000 & \textbf{1} \\
 & R vs F   & 0.1746 & 0.2878 & 0.9015 & 0.5996 & 0.1412 & 1.0000 & 1.0000 & 0 \\
\midrule
\multirow{5}{*}{Mistral-7B}
 & Initial  & 21.2 & 0.0 & 76.9 & 0.2  & 0.6  & 0.2 & 0.9 & \\
 & Revision & 20.8 & 0.0 & 77.1 & 0.0  & 2.0  & 0.0 & 0.1 & \\
 & Final    & 27.3 & 1.0 & 68.7 & 1.0  & 2.0  & 0.0 & 0.0 & \\
 & I vs R   & 0.9239 & 1.0000 & 1.0000 & 1.0000 & 0.1251 & 1.0000 & 1.0000 & 0 \\
 & R vs F   & 0.2030 & 0.2878 & 0.1313 & 0.2878 & 1.0000 & 1.0000 & 1.0000 & 0 \\
\midrule
\multirow{5}{*}{Mistral-24B}
 & Initial  & 2.2  & 0.0 & 11.0 & 46.9 & 39.8 & 0.0 &0.1 & \\
 & Revision & 10.2 & 0.4 & 33.6 & 30.3 & 25.4 & 0.0 & 0.1& \\
 & Final    & 5.1  & 0.0 & 14.1 & 42.4 & 38.4 & 0.0 &0.0 & \\
 & I vs R   & \textbf{0.0000} & 0.3333 & \textbf{0.0000} & \textbf{0.0000} & \textbf{0.0001} & 1.0000 & 1.0000 & \textbf{4} \\
 & R vs F   & 0.1437 & 1.0000 & \textbf{0.0003} & \textbf{0.0328} & \textbf{0.0184} & 1.0000 & 1.0000 & \textbf{3} \\
\midrule
\bottomrule
\end{tabular}
\caption{Feedback distribution (\%) and Fisher's Exact Test significance across draft stages under zero-shot prompting. \textbf{Bold} p-values indicate significant difference ($p < 0.05$). Sig. = number of feedback focus types with significant difference.}
\label{tab:fisher_drafts}
\end{table*} 

\textbf{RQ3: Do LLMs adapt their feedback focus based on student performance, as an expert teacher does?} To address this question, we examine whether feedback focus type distributions differ between high- and low-performing students for both the teacher and each model, using Fisher's Exact Test to detect significant differences, where significance suggests that feedback focus varies with student performance. This analysis is limited to the composition course, as it is the only course that provides essay grades with variation across students. For each draft stage, students are divided into high- and low-performing groups based on the median score (details in Appendix \ref{sec:Dataset}), and the proportion of each feedback type is computed for each group. 

As shown in Table~\ref{tab:fisher_performance} (zero-shot), the teacher's feedback shows significant differences between high- and low-performing students in Task Constraints and Praise. 
Gemini and Mistral-24B show significant differences in Clarification, LearnLM in Mistakes, and Qwen-70B in Elaboration, while Llama-7B and Mistral-7B show no significant differences across any feedback focus type. 
Except for Gemini, the feedback focus type that shows a significant difference corresponds to the type the model produces most frequently as in Figure~\ref{fig:distribution-feedback-focus}, suggesting that most models can only adapt to high- and low-performing students in feedback focus types they produce with sufficient frequency. Furthermore, all models that show adaptivity based on student performance are  large or closed models.

\begin{table*}[t]
\centering
\scriptsize
\setlength{\tabcolsep}{4pt}
\begin{tabular}{llrrrrrrrc}
\toprule
\textbf{Model/Teacher} & \textbf{Group} & 
\textbf{Task Constraints} & 
\textbf{Concepts} & 
\textbf{Elaboration} & 
\textbf{Clarification} & 
\textbf{Mistakes} & 
\textbf{Praise} & 
\textbf{Other} &
\textbf{Sig.} \\
\midrule
\multirow{3}{*}{Teacher}
& High    & 88  & 3 & 75  & 44  & 64  & 209 & 16 & \multirow{3}{*}{\textbf{2}} \\
& Low     & 101 & 1 & 59  & 31  & 48  & 91  & 4  & \\
& p-value & \textbf{$<$0.001}$^*$ & 0.653 & 0.337 & 0.902 & 0.536 & \textbf{$<$0.001}$^*$ & 0.068 & \\
\midrule
\multirow{3}{*}{Gemini}
& High    & 56  & --- & 286 & 37  & 120 & --- & --- & \multirow{3}{*}{1} \\
& Low     & 36  & --- & 182 & 41  & 76  & --- & --- & \\
& p-value & 0.910 & --- & 0.434 & \textbf{0.021}$^*$ & 0.678 & --- & --- & \\
\midrule
\multirow{3}{*}{LearnLM}
& High    & 99  & 0 & 288 & 36 & 76  & --- & --- & \multirow{3}{*}{1} \\
& Low     & 56  & 1 & 178 & 30 & 70  & --- & --- & \\
& p-value & 0.2765 & 0.4017 & 0.2009 & 0.3631 & \textbf{0.0407}$^*$ & --- & --- & \\
\midrule
\multirow{3}{*}{Qwen-70B}
& High    & 28  & 1 & 212 & 134 & 123 & --- & --- & \multirow{3}{*}{1} \\
& Low     & 22  & 1 & 112 & 110 & 90  & --- & --- & \\
& p-value & 0.6558 & 1.0000 & \textbf{0.0090}$^*$ & 0.0741 & 0.5172 & --- & --- & \\
\midrule
\multirow{3}{*}{Llama-7B}
& High    & 131 & 0   & 161 & 67  & 138 & ---  & 2  & \multirow{3}{*}{0} \\
& Low     & 79  & 2   & 106 & 53  & 94  & ---  & 1  & \\
& p-value & 0.416 & 0.161 & 0.880 & 0.365 & 0.937 & --- & 1.000 & \\
\midrule
\multirow{3}{*}{Mistral-7B}
& High    & 108 & 1 & 375 & 2   & 8   & 1   & 4  & \multirow{3}{*}{0} \\
& Low     & 74  & 0 & 259 & 0   & 2   & 0   & 0  & \\
& p-value & 0.932 & 1.000 & 0.508 & 0.519 & 0.331 & 1.000 & 0.153 & \\
\midrule
\multirow{3}{*}{Mistral-24B}
& High    & 30  & 1 & 99  & 190 & 179 & --- & --- & \multirow{3}{*}{1} \\
& Low     & 11  & 0 & 52  & 156 & 116 & --- & --- & \\
& p-value & 0.101 & 1.000 & 0.120 & \textbf{0.018}$^*$ & 0.712 & --- & --- & \\
\bottomrule
\end{tabular}
\caption{Feedback type counts for high and low performing students under zero-shot prompting. Fisher's Exact Test (p-values) was used to compare distributions between groups. $^*$ indicates a significant difference between high and low performing students  
$p < 0.05$. --- indicates zero counts in both groups 
(test skipped). Sig = number of feedback types with 
significant differences ($p < 0.05$).}
\label{tab:fisher_performance}
\end{table*}

\section{Conclusion}
We present the first systematic study of LLM alignment to expert teachers at the level of feedback focus type distributions in student writing. Using a content-based feedback focus taxonomy, we annotate teacher and LLM feedback across three writing courses and evaluate six LLMs under three prompting strategies. Our findings show that while most LLMs cover most feedback focus types, they fail to reflect teacher feedback distributions. In terms of adaptivity across draft stages, the teacher shows significant changes in five feedback focus types, while the best performing model shows significant changes in only three, and some models show no significant changes at all. Regarding student performance, some models show feedback  adaptivity between high and low performing students, but only in the feedback focus types they produce most frequently. To support future research, we release \textsc{FeedType}, a public corpus of annotated teacher and LLM feedback across three writing courses.

\section{Limitations}
Feedback adaptivity analysis is limited to the composition course taught by a single instructor. 
Future work should examine 
whether these patterns generalize to other writing contexts and instructors. Also, we focus on paragraph-level feedback, which 
is one common form of teacher feedback. 
Instructors also provide feedback at other 
levels, such as 
holistic 
essay-level feedback. Future research could 
examine how feedback 
distributions 
differ across these 
formats. 
While we evaluate three prompting strategies, 
fine-tuned models or models 
with explicit pedagogical instructions may 
show different alignment patterns. Future 
work could investigate whether targeted 
fine-tuning can elicit more adaptive and 
teacher-like LLM feedback. 

\section{Acknowledgements}
This work was supported by the Pitt Cyber Accelerator Grants Program. We extend our sincere gratitude to the Pitt PETAL group and 
anonymous reviewers for their insightful feedback.

\label{sec:Result Part}

\clearpage 

\bibliography{custom}
\clearpage      
\appendix

\clearpage      

\section{Dataset Overview}
\label{sec:Dataset}

\subsection{Data Selection}
The dataset consists of student essays across multiple 
draft stages (Initial, Revision, and Final), where each 
essay contains multiple paragraphs. Each paragraph is 
associated with one or more teacher feedback comments, 
and students received a numerical grade (0-100) for 
each essay assigned by the instructor.

\paragraph{Performance Groups.}
\label{sec:Performance Groups.}

We compute the median grade independently for each essay 
and draft stage to split students into high and low 
performing groups. Students at or above the median are 
classified as high performing and those below as low 
performing. Since the median is computed independently per essay and draft stage, the same student may be classified 
differently across drafts, reflecting their actual 
performance at each stage. The median grade and number 
of students in each group per essay and draft stage 
are reported in Table~\ref{tab:Distribution_selected_students}.

\paragraph{Paragraph Selection.}
\label{sec:Paragraph Selection.}
After computing the median split, we select only 
paragraphs that received exactly one feedback comment, 
as our focus is on paragraph-level feedback where each 
paragraph receives a single comment. Paragraphs with more than one comment are excluded, which results in excluding students who do not have any paragraphs with exactly one feedback comment. This leads to unequal counts of high and low performing students, as the median split is applied 
before paragraph selection. Table~\ref{tab:Distribution_selected_students} reports 
the number of selected students classified as high 
and low performing for each essay and draft stage.

\begin{table*}[t]
\centering
\scriptsize
\begin{tabular}{lcccccccc}
\toprule
Class & Essay & Draft & High ($n$) & Low ($n$) & Total ($n$) & Median &  Grade Range & Selected (High, Low) \\
\midrule

\multirow{13}{*}{Class A} 
& \multirow{3}{*}{Essay 1} & First    & 8 & 7 & 15 & 75 & 66-83 & 15 (8,7)\\  
&                          & Revision & 5 & 5 & 10 & 84 & 56-94 & 10 (5,5)\\
&                          & Final    & 3 & 2 & 5  & 85 & 75-93 & 3 (2,1)\\
& \multicolumn{2}{r}{\textbf{Essay 1 Total}} & 16 & 14 & 30 & - & - & 28 (15,13) \\
\cmidrule(lr){2-9}

& \multirow{3}{*}{Essay 2} & First    & 8 & 7 & 15 & 79 & 0-93 & 15 (8,7)\\  
&                          & Revision & 7 & 2 & 9  & 90 & 0-95 & 9 (7,2) \\
&                          & Final    & 3 & 3 & 6  & 93 & 68-97 & 0 \\
& \multicolumn{2}{r}{\textbf{Essay 2 Total}} & 18 & 12 & 30 & - & - & 24(15,9) \\
\cmidrule(lr){2-9}

& \multirow{3}{*}{Essay 3} & First    & 8 & 7 & 15 & 79 & 0-93 & 13 (7,6)\\  
&                          & Revision & - & - & - & - & - & - \\
&                          & Final    & 5 & 4 & 9  & 91 & 67-95 & 8 (5,3) \\
& \multicolumn{2}{r}{\textbf{Essay 3 Total}} & 13 & 11 & 24 & - & - & 21 (12,9) \\
\cmidrule(lr){2-9}
& \multicolumn{2}{r}{\textbf{Class A Total}} & 47 & 37 & 84 & - & - & 73 (42,31) \\

\midrule

\multirow{13}{*}{Class B} 
& \multirow{3}{*}{Essay 1} & First    & 8 & 7 & 15 & 73 & 61-86 & 15 (8,7)\\  
&                          & Revision & 9 & 5 & 14 & 86 & 61-93 & 13 (9,4) \\
&                          & Final    & 2 & 2 & 4  & 89 & 65-92 & 2 (2,0)\\
& \multicolumn{2}{r}{\textbf{Essay 1 Total}} & 19 & 14 & 33 & - & - & 30(19,11) \\
\cmidrule(lr){2-9}

& \multirow{3}{*}{Essay 2} & First    & 8 & 7 & 15 & 80 & 70-88 & 14 (8, 6) \\  
&                          & Revision & 8 & 5 & 13 & 93 & 77-95 & 10 (5,5)\\
&                          & Final    & 2 & 2 & 4  & 93 & 86-96 & 4 (2,2)\\
& \multicolumn{2}{r}{\textbf{Essay 2 Total}} & 18 & 14 & 32 & - & - & 28(15,13) \\
\cmidrule(lr){2-9}

& \multirow{3}{*}{Essay 3} & First    & 8 & 7 & 15 & 83 & 0-88 & 15 (8,7)\\  
&                          & Revision & - & - & - & - & - & - \\
&                          & Final    & 4 & 4 & 8 & 92.5 & 61-96 & 6 (3,3) \\
& \multicolumn{2}{r}{\textbf{Essay 3 Total}} & 12 & 11 & 23 & - & - & 21 (11,10) \\
\cmidrule(lr){2-9}
& \multicolumn{2}{r}{\textbf{Class B Total}} & 49 & 39 & 88 & - & - & 79 (45,34) \\
\midrule
\multicolumn{3}{l}{\textbf{Overall Total (Class A + B)}} & 96 & 76 & 172 & - & - & 152 (87,65) \\
\bottomrule
\end{tabular}
\caption{Distribution of selected student performance across classes, essays, and draft stages. High and low groups are defined using a median split. Grade range shows the lowest and highest grade per group. Totals are calculated per essay, per class, and overall.}
\label{tab:Distribution_selected_students}
\end{table*}

\subsection{Example Teacher Feedback in a Paragraph}
\label{sec:appendix_example}
Table~\ref{tab:example_paragraph} presents some examples of a student paragraph, with the teacher’s feedback and its annotation according to the feedback types.

\begin{table*}[t]
\centering
\small
\begin{tabular}{p{1.8cm} p{5.5cm} p{5.5cm} p{3cm}}
\toprule
\textbf{Draft Stage} & \textbf{Paragraph} & \textbf{Teacher Feedback} & \textbf{Feedback Type} \\
\midrule

\textbf{Initial} & My role as a woman is to bear. Bear, a verb in the Old English \textit{beran}, means “to carry, bring; bring forth, give birth to, produce; to endure, to wear.” According to etymonline.com, my role as a woman is to carry my husband’s child, produce an heir, and bring forth the next generation. My role as a woman is to endure the comments made by men as I walk by and to wear clothing that does not seduce men but not too modestly that I am considered boring. As a woman, my role is to bear. & This is a great beginning of a definition paragraph. What if you extended it? Got even more concrete and specific? Added layer upon layer from your own life and experience? & Elaboration \\

\addlinespace

\textbf{Revised} & We still get together for birthdays and important events my siblings and I may have. I learned how to be mature because I saw my parents having the right concept of being there for their children no matter what. It has been an important factor in my growth as a young woman because it taught me that this maturity, which is, in a way, a language, is what is going to help me develop relationships with other people and how to deal with difficult situations I may encounter with people along my life. & This is a fascinating and important paragraph, but I wonder if it's necessary for this essay and your essay's central focus. This is a genuine question. If it is necessary, perhaps it needs a little more development and connection. If not, what would be the effect if you simply cut it? & Task Constraints \\

\addlinespace

\textbf{Final} & Maybe writing with my grandfather’s pen brought comfort. The familiar hold, the familiar gold details, and the shiny black body all brought some sort of immediate ease and focus to my thoughts. & Nice. & Praise \\

\bottomrule
\end{tabular}
\caption{Examples of paragraphs across draft stages with teacher feedback and corresponding feedback type annotations.}
\label{tab:example_paragraph}
\end{table*}


\section{Narciss's Taxonomy and Our Adaptations}
\label{sec:narciss}

\citet{inbook2} proposes a content-based taxonomy of feedback components, distinguishing between simple and elaborated types. Simple feedback components include Knowledge of Performance (KP), Knowledge of Result (KR), and Knowledge of Correct Response (KCR). Elaborated feedback components include Knowledge about Task Constraints (KTC), Knowledge about Concepts (KC), Knowledge about Mistakes (KM), Knowledge about How to Proceed (KHP), and Knowledge about Metacognition (KMC). We adopt the elaborated feedback components as the basis for our feedback focus taxonomy, as they are more informative and better reflect the nature of writing feedback. Specifically, we retain Task Constraints, Concepts, and Mistakes, as these directly apply to writing tasks, mistakes in writing encompass a wide range of issues including grammar errors, organization, and coherence. We adapt KHP into \textit{Elaboration}, as in writing feedback, procedural guidance typically takes the form of asking students to develop and expand their ideas. We adapt KCR into \textit{Clarification}, as teachers often ask students to clarify their meaning rather than providing the correct response directly. We exclude the simple feedback components (KP, KR, KCR) as they are more suited to closed-ended tasks with clear correct answers, which does not apply to open-ended writing tasks. We also exclude KMC, as metacognitive feedback was rarely observed in our dataset. Finally, we add two categories not present in Narciss's taxonomy: \textit{Praise}, to capture positive reinforcement which is common in writing feedback, and \textit{Other}, to account for feedback instances that address multiple focus types simultaneously or do not fit any type.

\section{Annotation Guidelines}
\label{sec:annotation-guidelines}
We developed a coding framework informed by prior work on feedback analysis and patterns observed in our dataset. The framework categorizes the \textit{content} of feedback, the focus of what the feedback addresses, into seven types, summarized in Table~\ref{tab:feedback_taxonomy_full}.
\paragraph{Task Description}
Annotators assigned each feedback to the single category that best matched its content. For all categories, annotators followed the definitions provided in Table~\ref{tab:feedback_taxonomy_full}. When assigning labels such as \textit{Task Constraints} or \textit{Concepts}, annotators could also refer to the assignment rubric in addition to the category definitions. If a feedback item did not fit any category, or could reasonably belong to more than one category, annotators selected \textit{Other }.

\paragraph{Labeling Priorities.} 
When feedback could belong to two types, the following priorities apply:

\begin{itemize}
    \item \textit{Task Constraints} take priority over \textit{Mistakes}, as addressing assignment requirements is more important than correcting minor errors in idea organization or grammar.
    \item \textit{Clarification} take priority over \textit{Elaboration}, since resolving confusion is more critical than adding extra details.
    \item \textit{Other Categories} take priority over \textit{Praise}, because feedback that begins with encouragement but also addresses an issue should prioritize corrective content over positive feedback.
\end{itemize}

\begin{table*}[t]
\centering
\small
\begin{tabular}{p{3cm} p{7cm} p{5cm}}
\toprule
\textbf{Feedback Type} & \textbf{Definition} & \textbf{Example Feedback} \\
\midrule
\textbf{Task Constraints} & Clarifies assignment goals, required components, or evaluation criteria. & Is this the right language and tone for your audience?\\
\textbf{Concepts} & Explains or reinforces key concepts or terminology necessary for completing the task. & A narrative focuses on a single moment in time. \\
\textbf{Elaboration} & Encourages expanding or deepening ideas through additional detail, reflection, or development. & I'm interested! But let's dig a little deeper, here. Why, specifically, are you now comfortable? \\
\textbf{Clarification} & Signals difficulty understanding meaning, reference, or sequence and requests clarification. & what do you mean? \\
\textbf{Mistakes} & Identifies specific errors such as grammar, word choice, organization, or logical inconsistencies. & Watch out for sentence fragments. \\
\textbf{Praise} & Expresses encouragement or positive evaluation without suggesting changes. & Awesome template! Nice job with writing moves. \\
\textbf{Other } & Feedback that does not fit other categories, including personal reactions or symbols. & :) \\
\bottomrule
\end{tabular}
\caption{Feedback taxonomy with definitions and a concise example for each category.}
\label{tab:feedback_taxonomy_full}
\end{table*}

\section{Evaluating LLMs as Annotators}
\label{sec:Evaluating LLMs as Annotators}
Our framework incorporates LLMs as part of the automatic evaluation procedure. To ensure the reliability of the LLM annotator, we verify three requirements in our evaluation procedure. First, we assess the reliability of the annotation task itself by collecting both human and model annotations on a subset of the data and computing inter-annotator agreement. This step confirms that the taxonomy is well-defined and that annotators can reach reasonable agreement on the evaluation outcomes. Second, following~\citet{calderon-etal-2025-alternative}, we evaluate whether the LLM can serve as a reliable substitute for human annotators. Third, we conduct a validity analysis to ensure that conclusions drawn from the full set of LLM-generated annotations are consistent with those derived from human annotations. 

\subsection{Traditional Metrics}
\label{subsec:Evaluation_Reliability}

\paragraph{Gold-Standard Dataset.} 
\label{subsec:Gold-Standard Dataset.}
To compute inter-annotator agreement, we created a gold-standard reference dataset consisting of 100 feedback sampled from different essays. The sampled feedback covers multiple categories, with at least 10 examples per category, except for the Concepts category, which is less frequent in the dataset. The sample size was chosen to meet the evaluation of  Alt-Test (see Subsection~\ref{subsec:Evaluation_Alt-Test_1}), which requires at least 75 annotated instances to assess whether an alternative annotator can reliably replace human annotations. Three human annotators participated in labeling the reference dataset. Two annotators were involved in the initial development of the annotation guidelines, while a third annotator was added to satisfy the Alt-Test requirement of having at least three annotators. Before annotating the reference dataset, the third annotator was trained using the annotation guidelines, labeled 60 feedback from earlier pilot rounds, and resolved disagreements with one of the original annotators through discussion until consensus was reached. Agreement was evaluated on 40 new feedback selected from earlier pilot rounds, resulting in a Cohen’s Kappa of $\kappa = 0.68$. After this training, all three annotators annotated the full set of 100 feedback. After annotation, we used the human majority label for each feedback to calculate inter annotator agreement and accuracy. The majority label was defined as the label agreed upon by at least two annotators. If no agreement was reached, we followed the priority labeling rules described in Section~\ref{sec:annotation-guidelines}.
\paragraph{Reliability:} We assess the reliability of LLMs as annotators by computing inter annotator agreement using Cohen’s $\kappa$ between LLM predictions and human labels. This measures how closely LLM annotations align with human judgments beyond chance agreement \cite{sim2005kappa}. Using the gold standard sample described above, we prompted the candidate models to classify the feedback and computed inter-annotator agreement. Table~\ref{tab:overall_agreement} reports Cohen’s $\kappa$ agreement between each LLM annotator and the human majority vote, as well as agreement among human annotators. Gemini 3 Flash and GPT-4-Turbo show human-level alignment, outperforming LLaMA-3.1 8B and Qwen3-14B in the classification task.

\paragraph{Accuracy:} \label{subsec:Evaluation_Accuracy}
We evaluate LLM feedback annotation as a classification task and use accuracy to measure how often a model’s predictions match the ground truth for each feedback \cite{opitz2024closer}. Using the gold-standard sample described above, we prompted the candidate models to classify the feedback and computed accuracy. Table~\ref{tab:overall_agreement} reports accuracy for the LLMs.

\begin{table}[h!]
\centering
\scriptsize
\begin{tabular}{l l c c}
\toprule
\multicolumn{4}{c}{\textbf{LLM vs. Human Majority}} \\
\midrule
\textbf{Model} & \textbf{Examples} & \textbf{Accuracy} & \textbf{Cohen’s $\kappa$} \\
\midrule
Qwen      & 3-shot  & 0.60 & 0.51 \\
Qwen      & 7-shot  & 0.67 & 0.58 \\
Qwen      & 10-shot & 0.70 & 0.61 \\
\midrule
LLaMA     & 3-shot  & 0.55 & 0.42 \\
LLaMA     & 7-shot  & 0.71 & 0.71 \\
LLaMA     & 10-shot & 0.71 & 0.63 \\
\midrule
Gemini    & 3-shot  & 0.76 & 0.70 \\
Gemini    & 7-shot  & 0.81 & 0.75 \\
Gemini    & 10-shot & \textbf{0.82} & \textbf{0.77} \\
\midrule

GPT-4-Turbo & 3-shot  & 0.72 & 0.65 \\
GPT-4-Turbo & 7-shot  & 0.71 & 0.63 \\
GPT-4-Turbo & 10-shot & \textbf{0.80} & \textbf{0.74} \\
\midrule
\multicolumn{4}{c}{\textbf{Human vs. Human}} \\
\midrule
Human 1 vs. Human 2 & -- & -- & 0.72 \\
Human 1 vs. Human 3 & -- & -- & 0.68 \\
Human 2 vs. Human 3 & -- & -- & 0.70 \\
\bottomrule
\end{tabular}
\caption{Inter-annotator agreement across LLMs and humans. Accuracy reflects exact label agreement, while Cohen’s $\kappa$ measures chance-corrected agreement. The top section reports LLM agreement with the human majority label, and the bottom section reports pairwise agreement among human annotators.}
\label{tab:overall_agreement}
\end{table}

\begin{table}[h]
\centering
\scriptsize
\begin{tabular}{l l c}
\toprule
\textbf{Model} & \textbf{Examples} & \textbf{Winning Rate ($\omega$)} \\
\midrule
LLaMA         & 10-shot & 0.33 \\
LLaMA         & 7-shot  & 0.33 \\
LLaMA         & 3-shot  & 0.00 \\
\midrule
Qwen          & 10-shot & 0.00 \\
Qwen          & 7-shot  & 0.00 \\
Qwen          & 3-shot  & 0.00 \\
\midrule
Gemini        & 10-shot & \textbf{0.67} \\
Gemini        & 7-shot  & 0.33 \\
Gemini        & 3-shot  & 0.33 \\
\midrule
GPT-4-Turbo   & 10-shot & \textbf{0.67} \\
GPT-4-Turbo   & 7-shot  & 0.33 \\
GPT-4-Turbo   & 3-shot  & 0.33 \\
\midrule
\end{tabular}
\caption{Alt-Test results showing the winning rate ($\omega$) for each model across different few-shot settings, using $\epsilon = 0.1$. A winning rate greater than 0.5 indicates that the model outperforms the excluded human annotator in the majority of cases.}
\label{tab:alt-test-results-eps1}
\end{table}

\begin{table*}[t]
\centering
\small
\begin{tabular}{l l c c c c c c c}
\toprule
\textbf{Draft} & \textbf{ Annotator Type} &
\textbf{Task} &
\textbf{Concepts} &
\textbf{Elaboration} &
\textbf{Clarification} &
\textbf{Mistakes} &
\textbf{Praise} &
\textbf{Other } \\
\toprule

\multirow{2}{*}{Initial}
& Human annotator   & \textbf{23.8} & 0.5 & 23.4 & 14.5 & 11.2 & \textbf{25.2} & 1.4\\
& Gemini annotator  & \textbf{29.4} & 0.5 & 23.4 & 10.7 & 11.2 & \textbf{23.8} & 0.9 \\
\toprule

\multirow{2}{*}{Revision}
& Human annotator  & \textbf{12.5} & 0.6 & 10.0 & 7.5 & 23.8 & \textbf{41.9} & 3.8 \\
& Gemini annotator    & \textbf{12.5} & 0.6 & 8.1  & 7.5 & 25.0 & \textbf{41.9} & 4.4 \\
\toprule

\multirow{2}{*}{Final}
& Human annotator   & \textbf{3.7} & 0.0 & 0.0 & 3.7 & 11.1 & \textbf{77.8} & 3.7\\
& Gemini annotator    & \textbf{5.6} & 0.0 & 0.0 & 5.6 & 7.4  & \textbf{77.8} & 3.7\\
\toprule
\end{tabular}
\caption{Feedback category distribution (\%) across drafts for human and Gemini annotations on the set of 400 feedback.}
\label{tab:feedback_distribution_Validation}
\end{table*}

\begin{table*}[t]
\centering
\small
\begin{tabular}{llccccccc}
\toprule
\textbf{Performance} & \textbf{Annotator Type} &
\textbf{Task} &
\textbf{Concept} &
\textbf{Elaboration} &
\textbf{Clarification} &
\textbf{Mistake} &
\textbf{Praise} &
\textbf{Other } \\
\midrule

\multirow{2}{*}{High}
 & Human annotator  & 14.5 & 1.8 & 12.7 & 14.6& \textbf{16.4} & \textbf{30.9} & 9.1 \\
 & Gemini annotator & 19.0 & 0.6 & 15.3 & 6.8  & \textbf{15.3} & \textbf{40.5} & 2.5 \\
\midrule

\multirow{2}{*}{Low}
 & Human annotator  & 23.8 & 0.0 & 11.0 & 14.0 & \textbf{22.8} & \textbf{24.6} & 3.5 \\
 & Gemini annotator & 24.0 & 0.3 & 17.5 & 8.4 & \textbf{16.1} & \textbf{24.7} & 1.2 \\
\bottomrule
\end{tabular}

\caption{Feedback category distribution (\%) by student performance level (High \& Low) for human annotators and Gemini, on the set of 400 feedback.}
\label{tab:performance_feedback_comparison}
\end{table*}

\subsection{Alt-Test}
\label{subsec:Evaluation_Alt-Test_1}

The preceding group-level $\kappa$ evaluation does not determine whether a specific LLM is comparable to a human annotator at the individual level, nor does it offer a principled 
threshold for trusting the model’s annotations. As a result, it cannot justify replacing human annotation with an LLM. To address this limitation, we employ the Alt-Test, which compares the LLM’s annotations to those of individual human annotators rather than to the group as a whole. We evaluate each candidate LLM against each of the three human annotators individually, using our gold-standard feedback subset described in \ref{subsec:Gold-Standard Dataset.}. The test follows a leave-one-out approach, in which each human annotator is excluded in turn, and we evaluate whether the LLM’s annotations align more closely with the consensus of the remaining annotators than the excluded annotator does. For classification tasks, we use accuracy to measure alignment with the consensus. To account for the inherent variability and cost of human annotation, we introduce a tolerance margin $\epsilon$, allowing minor deviations, as implemented in the Alt-Test. Following prior work, $\epsilon$ is selected based on annotator expertise: $\epsilon = 0.2$ for experts, $\epsilon = 0.15$ for skilled annotators, and $\epsilon = 0.1$ for crowd workers. In our experiments, we observed that using larger $\epsilon$ values (e.g., 0.15 and 0.2) led the models to pass the test across all settings in Table~\ref{tab:overall_agreement}. Therefore, to enforce a stricter evaluation criterion, we adopt $\epsilon = 0.1$, reflecting the relatively low-expertise nature of the task. If the LLM aligns better than the excluded annotator in more than half of the cases (winning rate > 0.5), it is considered a viable replacement for human annotation. As shown in Table \ref{tab:alt-test-results-eps1}, the test yielded a winning rate of $\omega = 0.67$ for Gemini Flash and GPT-4-Turbo with 10 few-shot examples, indicating that the LLM’s judgments can be reliably trusted. This result supports the use of the LLM as an annotator in our evaluation framework. So, we select Gemini 3 Flash as the annotator due to its lower cost compared to GPT-4-Turbo.

\subsection{LLM Annotation Validity}
\label{subsec:Evaluations_Validation}

The prior metrics used in subsections ~\ref{subsec:Evaluation_Reliability} and~\ref{subsec:Evaluation_Alt-Test_1} do not guarantee that a model captures meaningful patterns in annotated data. As highlighted in prior work \cite{thomas2025beyond}, high agreement among annotators or strong performance on automated metrics does not necessarily reflect the validity of the  annotations. In educational contexts, annotation quality should be evaluated based on whether it captures patterns that are meaningful, interpretable, and relevant to learning \cite{thomas2025beyond}. Motivated by this perspective, we move beyond agreement based evaluation and assess whether the model reflects educationally relevant patterns in feedback. We therefore evaluate the validity of the LLM as an annotator by examining whether its annotations capture meaningful and educationally relevant patterns, and whether they align with those observed in human annotated data. To do this, we select a subset of the dataset designed to capture variation in student performance across draft stages. For each assignment and draft stage (initial, revision, and final), we calculate the median score across all students and use it to distinguish higher and lower performing students. From each assignment at each draft stage, we select two essays with the highest scores and two with the lowest scores. With three assignments and three draft stages per assignment, this results in a total of 36 essays from Class 1. For each selected essay, we extract the corresponding teacher feedback and annotate it according to the categories defined in our taxonomy, resulting in a total of 441 feedback annotations. This sampling strategy aims to identify meaningful patterns to answer two research questions and support evaluation. First, what types of feedback change (increase or decrease) between drafts. Second, what types of feedback change for high versus low performing students?

To address the first question, we examined the two most frequent feedback types, Praise and Task Constraints, that students received in the initial draft according to human annotations, tracked how their percentages changed across draft stages, and compared them with Gemini annotations for the same feedback types (see Table~\ref{tab:feedback_distribution_Validation}). Both human annotators and Gemini exhibited similar patterns: Praise feedback increased from the initial to final drafts, reflecting recognition of student improvements, while Task Constraint feedback decreased, indicating that students’ understanding of the assignment requirements improved across drafts.  For student performance, addressing the second question, we followed a similar approach as in the first question. We focused on the two most frequent feedback types received by high performing students according to human annotations, Praise and Mistakes, and examined how their percentages changed between high and low performing students, as well as in Gemini annotations (see Table~\ref{tab:performance_feedback_comparison}). Both human and Gemini annotations showed that Praise feedback decreased from high to low performing students, indicating that higher-performing students received more recognition. In contrast, Mistakes feedback increased, showing that lower performing students received more corrective feedback.
In both questions, Gemini was able to recognize the patterns identified by human annotations, as well as those reported in educational research on effective feedback practices as shown in Section~\ref{sec:Related Work}. In addition to capturing patterns observed in human annotations and the educational feedback literature, we compared the percentages of feedback categories between human annotations and Gemini. While the exact percentages were not always identical, Gemini generally reflected the trends seen in human annotations. Sometimes, however, Gemini was able to produce the same percentages. For example, the Praise category had very similar percentages in both the revision and final drafts, and categories such as Elaboration and Mistakes in the initial draft were also closely aligned.
\section{LLM Feedback}
\label{sec:prompt-llm-generators}
\subsection{Using LLMs for Feedback Generation}

We prompt the models under three conditions to generate feedback on student essay paragraphs. In all conditions, the model is provided with the entire essay, the target paragraph, the assignment rubric, and the draft stage. For revision and final drafts, previously received feedback is also included so the model can consider prior comments when generating updated feedback.
\noindent\textbf{Prompt 1 (Zero-shot):} The model receives no information about feedback focus types and is prompted to generate feedback directly.
\noindent\textbf{Prompt 2 (Zero-shot + Categories):} The model additionally receives the names and definitions of the seven feedback focus types to guide its output.
\noindent\textbf{Prompt 3 (Few-shot + Categories):} The model receives the names, definitions, and examples of each feedback focus type to further guide its output. Table~\ref{tab:prompt-conditions} shows the prompt conditions used for feedback generation.

\begin{table*}[t]
\centering
\scriptsize
\setlength{\tabcolsep}{3pt}
\renewcommand{\arraystretch}{0.92}
\begin{tabular}{p{0.25\textwidth}p{0.70\textwidth}}
\toprule
\textbf{Prompt} & \textbf{Content} \\
\midrule

\textbf{Prompt 1: Zero-shot} &
\textbf{\#\# Task Description} \newline
You are an experienced English teacher. Provide feedback for a student's \{draft\_stage\} essay draft using the \{previous\_feedback\_is\_available\} , current draft, target paragraph, and rubric. \newline
\textbf{\#\# Input} \newline
\textbf{Previous Feedback:} \{previous\_feedback\} \newline
\textbf{Current Draft:} \{current\_body\} \newline
\textbf{Target Paragraph:} \{paragraph\} \newline
\textbf{Rubric:} \{rubric\} \newline
\textbf{\#\# Your Task} \newline
1) Identify the SINGLE most important feedback focus in the TARGET PARAGRAPH. \newline
2) Write ONE short constructive feedback comment reflecting that focus. \newline
\textbf{\#\# Feedback Rules} \newline
- Feedback must be 1--4 sentences. \newline
- Focus on ONE feedback focus only. \newline
- Do NOT summarize or repeat the essay, list strengths and weaknesses, or add meta-commentary.

\textbf{\#\# Output Format} \newline
Feedback: <ONE paragraph of feedback>
\\
\midrule

\textbf{Prompt 2: Zero-shot + Categories} &
Same as Prompt 1, but the model also receives the names and definitions of the seven feedback focus types before the task instructions. \newline

\textbf{\#\# Feedback Focus Types} \newline
\textbf{Task Constraints:} Clarifies assignment goals, required components, or evaluation criteria. \newline
\textbf{Concepts:} Explains or reinforces key concepts or terminology necessary for completing the task. \newline
\textbf{Elaboration:} Encourages expanding or deepening ideas through additional detail, reflection, or development. \newline
\textbf{Clarification:} Signals difficulty understanding meaning, reference, or sequence and requests clarification. \newline
\textbf{Mistakes:} Identifies specific errors such as grammar, word choice, organization, or logical inconsistencies. \newline
\textbf{Praise:} Expresses encouragement or positive evaluation without suggesting changes. \newline
\textbf{Other:} Feedback that does not fit other categories or combines multiple feedback types. \newline

\\
\midrule

\textbf{Prompt 3: Few-shot + Categories} &
Same as Prompt 2, but the model also receives examples of each feedback focus type before the task instructions. \newline

\textbf{\#\# Feedback Focus Types} \newline
\{feedback\_focus\_definitions\} \newline
\textbf{\#\# Examples} \newline
\{feedback\_focus\_examples\}
\\
\midrule

\textbf{Feedback Classification} &
\textbf{\#\# Definitions} \newline
\textbf{Task Constraints:} Clarifies assignment goals, required components, or evaluation criteria. \newline
\textbf{Concepts:} Explains or reinforces key concepts or terminology necessary for completing the task. \newline
\textbf{Elaboration:} Encourages expanding or deepening ideas through additional detail, reflection, or development. \newline
\textbf{Clarification:} Signals difficulty understanding meaning, reference, or sequence and requests clarification. \newline
\textbf{Mistakes:} Identifies specific errors such as grammar, word choice, organization, or logical inconsistencies. \newline
\textbf{Praise:} Expresses encouragement or positive evaluation without suggesting changes. \newline
\textbf{Other:} Feedback that does not fit other categories or combines multiple feedback types. \newline

\textbf{\#\# Rubric} Use the rubric to guide classification. \newline
\textbf{\#\# Task} Classify the feedback into exactly one category. Output only the category name. \newline
\textbf{\#\# Example} Feedback: ``...'' Category: ``...'' \newline
\textbf{\#\# Target Feedback} \{feedback\_text\_here\}
\\
\bottomrule

\end{tabular}
\caption{Prompts used for feedback generation and feedback classification.}
\label{tab:prompt-conditions}
\end{table*}


\subsection{Examples of Feedback Generation}
\label{sec:Feedback Generation} 

Table~\ref{tab:human and model-generated feedback for a single paragraph.} presents a single paragraph along with feedback from multiple sources, including teacher and different models.

\begin{table*}[h]
\centering
\small
\begin{tabular}{p{6.5cm} p{3cm} p{6.5cm}}
\toprule
\textbf{Paragraph} & \textbf{Source} & \textbf{Feedback and its type} \\
\midrule

\multirow{5}{6.5cm}{
My brain uses up all my energy. All I do is overthink, overthink, and overthink. I spend every inch of my head over analyzing everything. All the possibilities and outcomes of every path that a choice opens. I will go over the past and the future. I exhaust myself rather than act and be able to work. Like the lamp, we have the ability to control the dimness of the light to be able to decide how much energy we want to use. So, our judgments shouldn't take a toll on us and should also take in less energy so more energy can go towards taking a step forward and allow us to physically perform.
} 
& Teacher 
& This is an interesting moment of friction that we could discuss! Let's see if we can find your essay's central focusing idea. Then, let's "map" this essay so that you know which parts of it are necessary, which aren't, and which you still need to develop. 

\textbf{Type of Feedback: Elaboration} \\
\cmidrule(lr){2-3}
& Mistral 7B 
& Your essay effectively explores the relationship between your mental energy expenditure and the metaphor of a lamp's adjustable light. However, consider expanding upon the connection between your personal experiences and the broader implications for readers. Delve deeper into how controlling the dimness of your thoughts parallels the physical control of a lamp's light. 

\textbf{Type of Feedback: Elaboration}\\
\cmidrule(lr){2-3}
& LLaMA 3-8B 
& In this paragraph, consider varying sentence structure to maintain a consistent pace and flow. You tend to start each sentence with a similar phrase ("I", "Like"), creating a sense of repetition that disrupts the overall rhythm of the text. Experiment with starting sentences with different phrases or transitioning smoothly between clauses to enhance the clarity and engagement of your argument.

\textbf{Type of Feedback: Mistakes}\\
\cmidrule(lr){2-3}
& Mistral 24B 
& You have used repetition effectively to emphasize your point about overthinking, but consider varying your sentence structure to maintain reader engagement. 

\textbf{Type of Feedback: Mistakes}\\
\cmidrule(lr){2-3}
& Gemini 2 Flash 
& You're making an interesting connection between your overthinking and the lamp's dimmable light, but the paragraph feels a bit repetitive and abstract. To strengthen your argument, try grounding this comparison in a specific example of a time when your overthinking (or lack thereof) directly impacted a decision you made. Showing the reader this connection through a concrete scenario will make your point more vivid and persuasive. 

\textbf{Type of Feedback: Elaboration}\\
\bottomrule
\end{tabular}
\caption{Human and model generated feedback for a single paragraph.}
\label{tab:human and model-generated feedback for a single paragraph.}
\end{table*}

\section{Using an LLM as an Annotator}
\label{sec:prompt-llm-judges}
We prompt the models in both zero shot and few shot settings to evaluate feedback on student essay paragraphs by classifying it into one of seven predefined categories. The prompt includes the list of these seven categories along with their definitions and rubric, as well as the target paragraph and the corresponding feedback to be annotated. For the few shot setting, we construct examples using newly selected data from Class 1, annotated by the first author. Specifically, we include at least 10 examples for each feedback category, except for the Concept category, which is relatively rare in our dataset. In total, 85 annotated feedback instances are used as examples to guide the annotation task. Table~\ref{tab:prompt-conditions} shows the prompt conditions used for feedback classification.

\end{document}